\def\year{2019}\relax
\documentclass[letterpaper]{article} %DO NOT CHANGE THIS
\usepackage{aaai19}  %Required
\usepackage{times}  %Required
\usepackage{helvet}  %Required
\usepackage{courier}  %Required
\usepackage{url}  %Required
\usepackage{graphicx}  %Required
\usepackage{subfigure}
\usepackage{amsfonts}
\usepackage{amssymb}
\usepackage{amsmath}

\begin{document}
\title{TET-GAN: Text Effects Transfer via Stylization and Destylization}
\author{Shuai Yang,~Jiaying Liu\thanks{Corresponding author. This work was supported by National Natural Science Foundation of China under contract No. 61772043 and Peking University - Tencent Rhino Bird Innovation Fund.\protect\\
Code and dataset will be available at \protect\url{http://www.icst.pku.edu.cn/struct/Projects/TETGAN.html}
},~Wenjing Wang~and~Zongming Guo\\
Institute of Computer Science and Technology, Peking University, Beijing, China\\
\{williamyang, liujiaying, daooshee, guozongming\}@pku.edu.cn}
\maketitle

% 传统方法基于缺乏效率的图像块，对文字形状的变化不鲁棒。
% 传统方法存在问题
% 两个task
% one-shot
% 为了训练网络，我们提出了XX数据集。
% 显式的表示，使得通过我们的网络架构设计，TET-GAN能够学习在这两个任务中处理所有这些字效，
% 我们的灵活架构设计，同时，只需要一张风格图片，通过one-shot  finetuning就能扩展到新的字效。
\begin{abstract}
Text effects transfer technology automatically makes the text dramatically more impressive. However, previous style transfer methods either study the model for general style, which cannot handle the highly-structured text effects along the glyph,
or require manual design of subtle matching criteria for text effects.
In this paper, we focus on the use of the powerful representation abilities of deep neural features for text effects transfer.
For this purpose, we propose a novel Texture Effects Transfer GAN (TET-GAN), which consists of a stylization subnetwork and a destylization subnetwork.
The key idea is to train our network to accomplish both the objective of style transfer and style removal, so that it can learn to disentangle and recombine the content and style features of text effects images.
To support the training of our network, we propose a new text effects dataset with as much as 64 professionally designed styles on 837 characters.
We show that the disentangled feature representations enable us to transfer or remove all these styles on arbitrary glyphs using one network.
Furthermore, the flexible network design empowers TET-GAN to efficiently extend to a new text style via one-shot learning where only one example is required.
We demonstrate the superiority of the proposed method in generating high-quality stylized text over the state-of-the-art methods.
\end{abstract}

%%%%%%%%% BODY TEXT
\section{Introduction}

Text effects are additional style features for text, such as colors, outlines, shadows, reflections, glows and textures. Rendering text in the style specified by the example stylized text is referred to as text effects transfer.
Applying visual effects to text is very common yet important in graphic design.
However, manually rendering text effects is labor intensive and requires great skills beyond normal users.
In this work, we propose a neural network architecture that automatically synthesizes high-quality text effects on arbitrary glyphs.

\begin{figure}[t]
  \centering
    \includegraphics[width=0.93\linewidth]{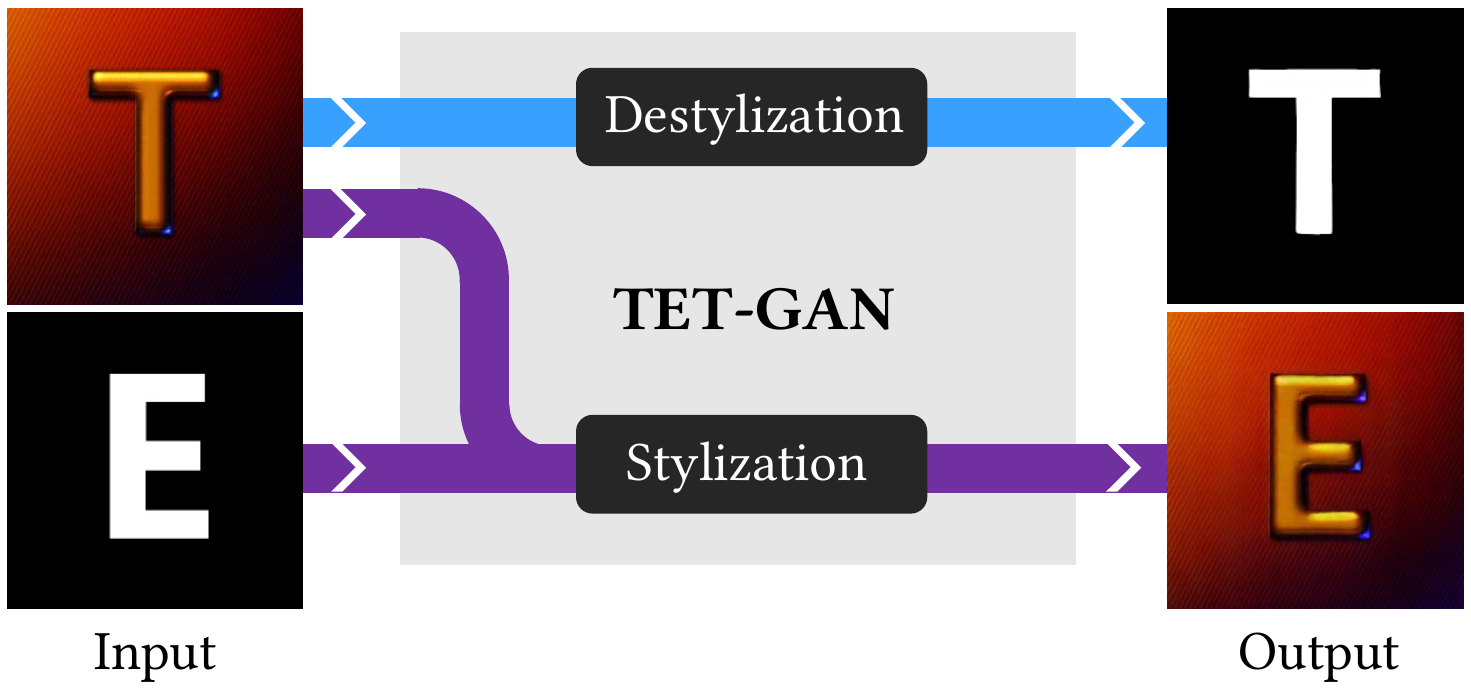}\vspace{-2mm}
  \caption{Overview: Our TET-GAN implements two functions: destylization for removing style features from the text and stylization for transferring the visual effects from highly stylized text onto other glyphs.}\label{fig:overview}\vspace{-2mm}
\end{figure}

%What did recent efforts do. What is their limitations.
% 1. Gram-based和那些IN、BN等全局统计调整方法不符合字效的风格。字效具有高度的结构化特征，不能简单将其描述为一个全局的纹理之间的相关度。而应该文字的结构生成，为此我们考虑数据驱动，制作了成对的训练数据，让网络自己学会如何对字效进行重构。！！！
The success of the pioneering Neural Style Transfer~\cite{gatys2016image} has sparked a research boom of deep-based image stylization.
%~\cite{gatys2016image,Johnson2016Perceptual,ulyanov2016texture,Wang2016Multimodal,Gatys2017Controlling,Chen2017StyleBank,Li2017Diversified}.
The key idea behind it is to match the global feature distributions between the style image and the generated image~\cite{Li2017Demystifying} by minimizing the difference of Gram matrices~\cite{gatys2015texture}. However, this global statistics representation for general styles does not apply to the text effects. Text effects are highly structured along the glyph and cannot be simply characterized as the mean, variance or other global statistics~\cite{Li2017Demystifying,Dumoulin2016A,huang2017adain,WCT-NIPS-2017} of the texture features. Instead, the effects should be learned with the corresponding glyphs. For this reason, we develop a new text effects dataset, driven by which we show our network can learn to properly rearrange textures to fit new glyphs.

% 2. patch-based。以重构的角度来看，patch-based似乎更合适，CVPR17无法处理文字变化巨大的情况；patchswap、CNNMRF、Neural Doodles，没有考虑到距离的引导，同时，黑白map 难以提取feature map，难以用来匹配。我们加入了距离的先验。
From a perspective of texture rearrangement, modelling the style of text effects using local patches seems to be more suitable than global statistics. Valuable efforts have been devoted to patch-based style transfer~\cite{Li2016Combining,Chen2016Fast,Yang2017Awesome}. The recent work of~\cite{Yang2017Awesome} is the first study of text effects transfer, where textures are rearranged to correlated positions on text skeleton. However, matching patches in the pixel domain,
this method fails to find proper patches if the target and example glyphs differ a lot. The recent deep-based methods~\cite{Li2016Combining,Chen2016Fast} address this issue by matching glyphs in the feature domain, but they use a greedy optimization, causing the disorder of the texture in the spatial distribution. To solve this problem, we introduce a novel distribution-aware data augmentation strategy to constrain the spatial distribution of textures.

% 3. I2I: 一个网络只能生成一种style，we introduce a set of problems where the forward and backward functions are asymmetric， CycleGAN could in principle learn to apply a general make-you-look-good makeup to a no-makeup face, but it would not replicate a specific example makeup style.；StarGAN能处理多种预先定义好的风格。但是当只有少量字效图，甚至只有一对图像的时候，无法扩展。我们的方法采用字形字效分离再组合的方式，是的一个网络可以处理64种风格。同时，支持few-shot的训练，在finetune，就可以处理全新的字效。% 4. 我们的方法在用字形字效分离的方法，可以完成风格化，去风格化，风格交换的功能。带来的好处在于，对于一些字效，我们甚至可以完成unsupervised style transfer。我们先去风格化，再finetune。
To handle a particular style, researchers have looked at style modeling from images rather than using  general statistics or patches, which refers to image-to-image translation~\cite{Isola2017Image}. Early attempts~\cite{Isola2017Image,Zhu2017Unpaired} train generative adversarial networks (GAN) to map images from two domains, which is limited to only two styles. StarGAN~\cite{Choi2017StarGAN} employs one-hot vectors to handle multiple pre-defined styles, but requires expensive data collection and retraining to handle new styles.
%cannot expand to unseen styles without expensive data collection and retraining.
We improve them by designing a novel Texture Effects Transfer GAN (TET-GAN), which characterizes glyphs and styles separately. By disentangling and recombining glyph and visual effects features, we show our network can simultaneously support stylization and destylization on a variety of text effects as shown in Fig.~\ref{fig:overview}. In addition, having learned to rearrange textures based on glyphs, a trained network can easily be extended to new user-specified text effects.

% 去风格化任务能够引导网络完全去除风格，提取内容，从而风格化网络能够更好地学会风格和内容的融合方法。
In this paper, we propose a novel approach for text effects transfer with three distinctive aspects.
First, we develop a novel TET-GAN built upon encoder-decoder architectures.
The encoders are trained to disentangle content and style features in the text effects images.
Stylization is implemented by recombining these two features while destylization by solely decoding content features. %The disentangled representations bridge the destylization and stylization subnetworks, enabling a joint training to improve each other.
The task of destylization to completely remove styles guides the network to precisely extract the content feature, which in turn helps the network better capture its spatial relationship with the style feature in the task of stylization.
Second, in terms of data, we develop a new text effects dataset with 53,568 image pairs to facilitate training and further study.
In addition, we propose a distribution-aware data augmentation strategy to impose a distribution constraint~\cite{Yang2017Awesome} for text effects.
Driven by the data, our network learns to rearrange visual effects according to the glyph structure and its correlated position on the glyph as a professional designer does.
Finally, we propose a self-stylization training scheme for one-shot learning. Leveraging the skills that have been learned from our dataset, the network only needs to additionally learn to reconstruct the texture details of one example, and then it can generate the new style on any glyph.
% 提出一个新颖的disentangled representation framework字效迁移框架，能够进行分离字形和字效，Due to the explicit style representation, our method enables 用一个网络完成多种字效的风格化和去风格化的任务。
% 设计了一个finetune的方案，使之支持few-shot的training，Our method not only allows to simultaneously train multiple styles sharing a single auto-encoder, but also incrementally learn a new style without changing the auto-encoder.
% 搜集了训练集，并透过数据预处理的方式，引入了距离先验

In summary, our contributions are threefold:
\begin{itemize}
  \item We raise a novel TET-GAN to disentangle and recombine glyphs and visual effects for text effects transfer. The explicit content and style representations enable effective stylization and destylization on multiple text effects.
  \item We introduce a new dataset containing thousands of professionally designed text effects images, and
      propose a distribution-aware data augmentation strategy for distribution-aware style transfer.
  \item We propose a novel self-stylization training scheme that requires only a few or even one example to learn a new style upon a trained network.
\end{itemize}

\begin{figure*}[htbp]
  \centering
  \includegraphics[width=1\linewidth]{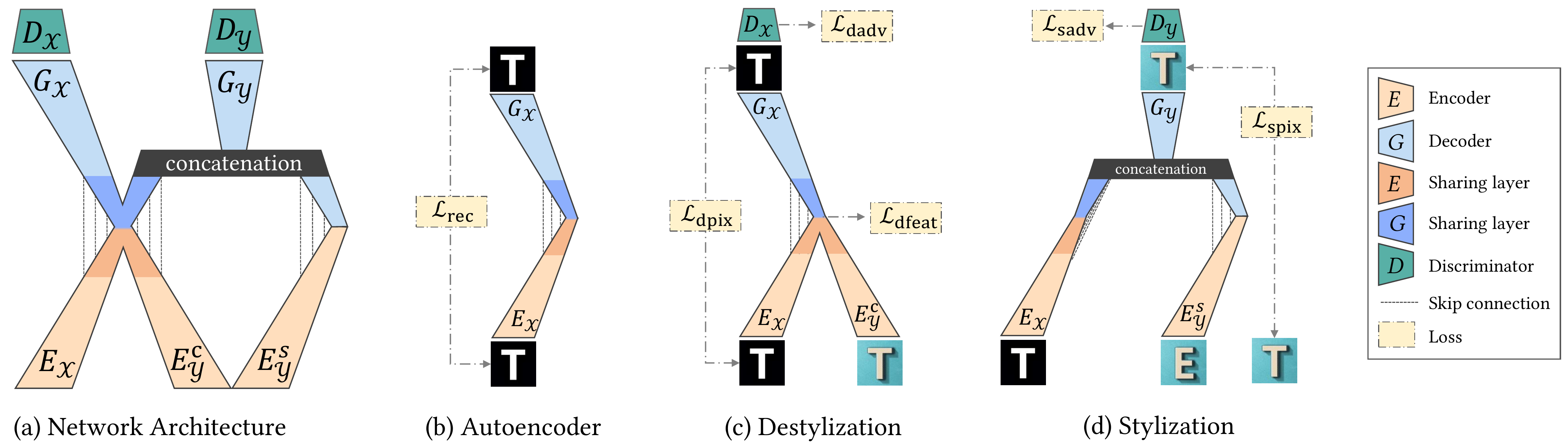}
  \caption{The TET-GAN architecture. (a) An overview of TET-GAN architecture. Our network is trained via three objectives of autoencoder, destylization and stylization. (b) Glyph autoencoder to learn content features. (c) Destylization by disentangling content features from text effect images. (d) Stylization by combining content and style features. }\label{fig:network-overview}
\end{figure*}

%-------------------------------------------------------------------------
\section{Related Work}

\textbf{Neural style transfer.}
Style transfer is the task of migrating styles from an example style image to a content image, which is closely related to texture synthesis.
%Early work models textures as local pixel intensity information, and focus on the image pixel or patch synthesis~\cite{Hertzmann2001Image,Efros2001Image}.
The pioneering work of~\cite{gatys2016image} demonstrates the powerful representation ability of convolutional neural networks to model textures. Gatys~\textit{et al.} formulated textures as the correlation of deep features in the form of a Gram matrix~\cite{gatys2015texture}, and transferred styles by matching high-level representations of the content image and the Gram matrices.
%, which is essentially a match of global feature distributions~\cite{Li2017Demystifying}.
Since then, deep-based style transfer has become a hot topic, and many follow-up work improves it in different aspects such as acceleration~\cite{Johnson2016Perceptual,ulyanov2016texture,Wang2016Multimodal}, user controls~\cite{Gatys2017Controlling} and style diversification~\cite{Li2017Diversified}.
In parallel, Li~\textit{et al.}~\shortcite{Li2016Combining,Li2016Precomputed} modelled textures by local patches of feature maps, which can transfer photo-realistic styles.
% 这个方法被证明是XXX，平行地，XXX

\textbf{Image-to-image translation.}
Image-to-image translation is a domain transfer problem, where the input and output are both images.
Driven by the great advances of GAN, once been introduced by~\cite{Isola2017Image}, it has been widely studied. Recent work~\cite{Murez2017Image} has been able to generate very high-resolution photo-realistic images from semantic label maps. Zhu~\textit{et al.}~\shortcite{Zhu2017Unpaired} proposed a novel cycle loss to learn the domain translation without paired input-output examples. While most researches focus on the translation between two domains, Choi~\textit{et al.}~\shortcite{Choi2017StarGAN} utilized a one-hot vector to specify the target domain, so that the network can learn the mapping between multiple domains, which provides more flexibility. However, extension to new domains is still expensive. In this paper, we introduce a self-stylization training scheme to efficiently learn a new style with only one example required.

\textbf{Text style transfer.}
Text is one of the most important visual elements in our daily life and there is some work on style transfer specific to the text. Taking advantage of the accessibility of abundant font images, many works~\cite{Lian2016Automatic,Sun2017Learning,Zhang2017Separating} trained neural networks to learn stroke styles for font transfer. However, another type of style, namely text effects, was not studied much.
It was not until 2017 that the work of \cite{Yang2017Awesome} first raised text effects transfer problem.
The authors proposed to match and synthesize image patches based on their correlated position on the glyph, which is vulnerable to glyph differences and has a heavy computational burden. Meanwhile, Azadi~\textit{et al.}~\shortcite{Azadi2017Multi} combined font transfer and text effects transfer using two successive subnetworks and end-to-end trained them using a synthesized gradient font dataset. However, they can only handle $26$ capital letters with a small size of $64\times64$, and their synthesized dataset differs greatly from the actual text effects. By contrast, we build our dataset using in-the-wild text effects with a size of $320\times320$, supporting our network to render exquisite text effects for any glyph.

\section{TET-GAN for Text Effects Transfer}

% 定义问题
Our goal is to learn a two-way mapping between two domains $\mathcal{X}$ and $\mathcal{Y}$, which represent a collection of text images and text effects images, respectively.
% 定义网络构造 E G D
Our key idea is to train a network to simultaneously accomplish two tasks: one to combine text effects (style) with glyphs (content) for stylization, and another to remove text effects for destylization.
As shown in Fig.~\ref{fig:network-overview}, our framework consists of two content encoders $\{E_{\mathcal{X}},E^c_{\mathcal{Y}}\}$,
a style encoder $\{E^s_{\mathcal{Y}}\}$,
two domain generators $\{G_{\mathcal{X}},G_{\mathcal{Y}}\}$ and
two domain discriminators $\{D_{\mathcal{X}},D_{\mathcal{Y}}\}$.
% 定义作用
% EX和Esc将X和Y映射到共享的字形隐空间，EYs将Y映射到字效隐空间，GX根据字形特征恢复文字图像，GY 根据字形特征和字效特征恢复字效图像。DX和DY 分别区分GX和GY生成图像的真假。
$E_{\mathcal{X}}$ and $E^c_{\mathcal{Y}}$ map text images and text effects images onto a shared content feature space, respectively, while $E^s_{\mathcal{Y}}$ maps text effects images onto a style feature space.
$G_{\mathcal{X}}$ generates text images from the encoded content features. $G_{\mathcal{Y}}$ generates text effects images conditioned on both the encoded content features and style features.
The discriminators are trained to distinguish the generated images from the real ones.
% 定义任务
% 1. AE：
% 2. DESTY:
% 3. STY:

Given these basic network components, we can define our two-way mapping.
The forward mapping (stylization) $G_{\mathcal{Y}}\circ(E_{\mathcal{X}},E^s_{\mathcal{Y}}):\mathcal{X}\times\mathcal{Y}\rightarrow\mathcal{Y}$ requires a target raw text image $x$ and an example text effects image $y'$ as input, and transfers the text effects in $y'$ onto $x$, obtaining $y$.
Meanwhile, the backward mapping (destylization) $G_{\mathcal{X}}\circ E^c_{\mathcal{Y}}:\mathcal{Y}\rightarrow\mathcal{X}$ takes $y$ as the input, and extracts its corresponding raw text image $x$. In addition, we further consider an autoencoder $G_{\mathcal{X}}\circ E_{\mathcal{X}}:\mathcal{X}\rightarrow\mathcal{X}$, which helps guide the training of destylization.
Our objective is to solve the min-max problem:
\begin{equation}\label{eq:total_loss}
  \min_{E,G}\max_{D} \mathcal{L}_{\text{rec}}+\mathcal{L}_{\text{desty}}+\mathcal{L}_{\text{sty}},
\end{equation}
where $\mathcal{L}_{\text{rec}}$, $\mathcal{L}_{\text{desty}}$, and $\mathcal{L}_{\text{sty}}$ are loss functions related to the autoencoder reconstruction, destylization, and stylization, respectively.
In the following sections, we present the detail of the loss functions and introduce our one-shot learning strategy that enables the training with only one example.

\subsection{Autoencoder}
\label{sec:ae}
% 1. AE：为了让字形特征保留足够的字形信息，我们要求字形特征能重建原文字图像
\textbf{Reconstruction loss}. First of all, the encoded content feature is required to preserve the core information of the glyph.
Therefore, we impose a reconstruction constraint that forces the content feature to completely reconstruct the input text image, leading to the standard autoencoder $L_1$ loss:
\begin{equation}\label{eq:ae_rec_loss}
  \mathcal{L}_{\text{rec}}=\lambda_{\text{rec}}\mathbb{E}_{x}[\|G_{\mathcal{X}}(E_{\mathcal{X}}(x))-x\|_{1}].
\end{equation}

\subsection{Destylization}
\label{sec:desty}
% 2. DESTY: 为了让XXX共同空间，sharing weight。通过pix和feature，feature相当于是Guidance，让其完全去除sty的成分，
In the training of our destylization subnetwork, we sample from the training set a text-style pair $(x,y)$.
We would like to map $x$ and $y$ onto a shared content feature space, where the feature can be used to reconstruct $x$. To achieve it, we apply two strategies: weight sharing and content feature guidance. First, as adopted in other domain transfer works~\cite{Liu2017Unsupervised,Murez2017Image}, the weights between the last few layers of $E_{\mathcal{X}}$ and $E^{c}_{\mathcal{Y}}$ as well as the first few layers of $G_{\mathcal{X}}$ and $G_{\mathcal{Y}}$ are shared. Second, we propose a feature loss to guide $E^{c}_{\mathcal{Y}}$ using the content feature extracted by the autoencoder. The total loss takes the following form:
% 除此之外，我们采用标准的cGAN框架，包含pix loss和adversarial lovv，因此我们的loss具有如下形式：
\begin{equation}\label{eq:desty_total_loss}
  \mathcal{L}_{\text{desty}}=\lambda_{\text{dfeat}}\mathcal{L}_{\text{dfeat}}+
  \lambda_{\text{dpix}}\mathcal{L}_{\text{dpix}}+\lambda_{\text{dadv}}\mathcal{L}_{\text{dadv}},
\end{equation}
where $\mathcal{L}_{\text{dfeat}}$ is the feature loss. Following the image-to-image GAN framework~\cite{Isola2017Image}, $\mathcal{L}_{\text{dfeat}}$ and $\mathcal{L}_{\text{dadv}}$ are pixel and adversarial losses, respectively.

\textbf{Feature loss}. The content encoder is tasked to approach the ground truth content feature. Let $S_{\mathcal{X}}$ denote the sharing layers of $G_{\mathcal{X}}$.
Then the content feature for guidance is defined as $z=S_{\mathcal{X}}(E_{\mathcal{X}}(x))$ and our feature loss is:
\begin{equation}\label{eq:desty_feat_loss}
  \mathcal{L}_{\text{dfeat}}=\mathbb{E}_{x,y}[\|S_{\mathcal{X}}(E^{c}_{\mathcal{Y}}(y))-z\|_{1}].
\end{equation}
Our feature loss guides the content encoder $E^c_{\mathcal{Y}}$ to remove the style elements from the text effects image, preserving only the core glyph information.

\textbf{Pixel loss}. The destylization subnetwork is tasked to approach the ground truth output in an $L_1$ sense:
\begin{equation}\label{eq:desty_pix_loss}
  \mathcal{L}_{\text{dpix}}=\mathbb{E}_{x,y}[\|G_{\mathcal{X}}(E^{c}_{\mathcal{Y}}(y))-x\|_{1}].
\end{equation}

\textbf{Adversarial loss}. We impose conditional adversarial loss to imporve the quality of the generated results. We adopt a conditional version of WGAN-GP~\cite{Gulrajani2017Improved} as our loss function,
where $D_{\mathcal{X}}$ learns to determine the authenticity of the input text image and whether it matches the given text effects image. At the same time, $G_{\mathcal{X}}$ and $E^{c}_{\mathcal{Y}}$ learn to confuse $D_{\mathcal{X}}$:
\begin{equation}\label{eq:desty_gan_loss}
\begin{aligned}
  \mathcal{L}_{\text{dadv}}=&\mathbb{E}_{x,y}[D_{\mathcal{X}}(x,y)]-\mathbb{E}_{y}[D_{\mathcal{X}}(G_{\mathcal{X}}(E^{c}_{\mathcal{Y}}(y)),y)]\\
  -&\lambda_{gp}\mathbb{E}_{\hat{x},y}[(\|\nabla_{\hat{x}}D_{\mathcal{X}}(\hat{x},y)\|_2-1)^2],
\end{aligned}
\end{equation}
where $\hat{x}$ is defined as a uniformly sampling along the straight line between the sampled real data $x$ and the sampled generated data $G_{\mathcal{X}}(E^{c}_{\mathcal{Y}}(y))$.

\begin{figure}[t]
  \centering
  \subfigure[]{
  \includegraphics[width=0.31\linewidth]{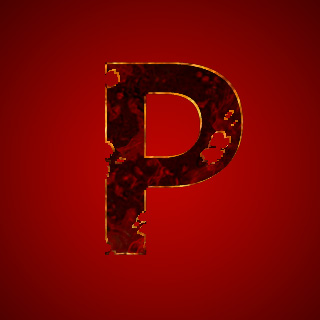}}
  \subfigure[]{
  \includegraphics[width=0.31\linewidth]{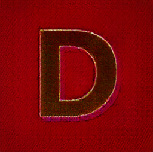}}
  \subfigure[]{
  \includegraphics[width=0.31\linewidth]{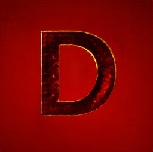}}
  \subfigure[self-stylization training scheme]{
  \includegraphics[width=0.96\linewidth]{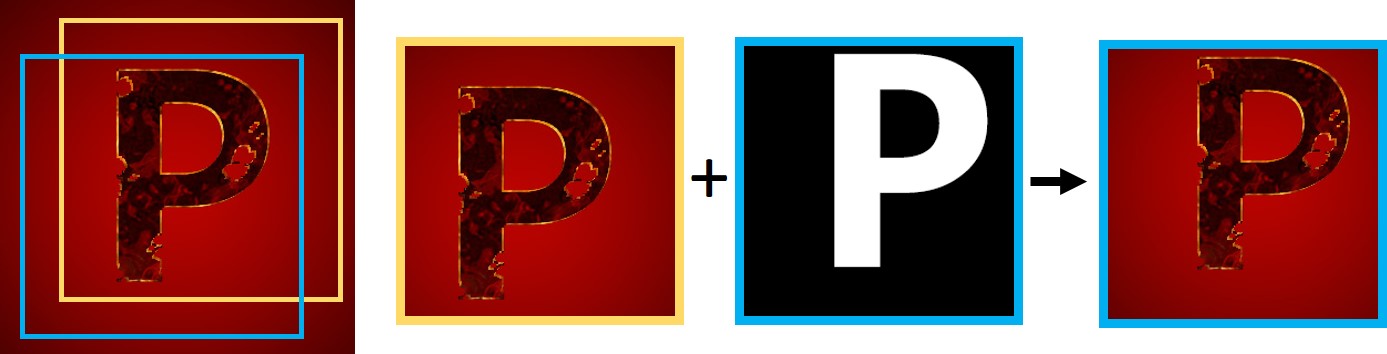}}
  \caption{One-shot text effects transfer. (a) User-specified new text effects.
  (b) Stylization result on an unseen style. (c) Stylization result after one-shot finetuning.
  (d) Randomly crop the style image to generate image pairs for training.}\label{fig:finetune}
\end{figure}

\subsection{Stylization}
\label{sec:sty}

% 3. STY: 我们首先提取字形和字效特征，将其串联后，进入到G网络中生成字效图。
For the stylization subnetwork, we sample from the training set a text-style pair $(x,y)$ and a text effects image $y'$ that shares the same style with $y$ but has a different glyph.
We first extract the content feature from $x$ and the style feature from $y$, which are then concatenated and fed into $G_{\mathcal{Y}}$ to generate a text effects image to approximate to the ground truth $y$. The standard image-to-image GAN loss is used:
\begin{equation}\label{eq:sty_total_loss}
  \mathcal{L}_{\text{sty}}=\lambda_{\text{spix}}\mathcal{L}_{\text{spix}}+\lambda_{\text{sadv}}\mathcal{L}_{\text{sadv}}.
\end{equation}

\textbf{Pixel loss}. The stylization subnetwork is tasked to approach the ground truth output in an $L_1$ sense:
\begin{equation}\label{eq:sty_pix_loss}
  \mathcal{L}_{\text{spix}}=\mathbb{E}_{x,y,y'}[\|G_{\mathcal{Y}}(E_{\mathcal{X}}(x),E^{s}_{\mathcal{Y}}(y'))-y\|_{1}].
\end{equation}

\textbf{Adversarial loss}. Similar to the destylization subnetwork, WGAN-GP~\cite{Gulrajani2017Improved} is employed where the discriminator's decision is conditioned by both $x$ and $y'$:
\begin{equation}\label{eq:sty_adv_loss}
\begin{aligned}
  \mathcal{L}_{\text{sadv}}=&\mathbb{E}_{x,y,y'}[D_{\mathcal{Y}}(x,y,y')]\\
  -&\mathbb{E}_{x,y'}[D_{\mathcal{Y}}(x,G_{\mathcal{Y}}(E_{\mathcal{X}}(x),E^{s}_{\mathcal{Y}}(y')),y')]\\
  -&\lambda_{gp}\mathbb{E}_{x,\hat{y},y'}[(\|\nabla_{\hat{y}}D_{\mathcal{Y}}(x,\hat{y},y')\|_2-1)^2].
\end{aligned}
\end{equation}
In the above equation, $\hat{y}$ is similarly defined as $\hat{x}$ in Eq.~(\ref{eq:desty_gan_loss}).
Note that we do not impose any other constraints (for example, style autoencoder reconstruction: $\mathcal{Y}\rightarrow\mathcal{Y}$ and cycle consistency: $\mathcal{X}\times\mathcal{Y}\rightarrow\mathcal{Y}\rightarrow\mathcal{X}$) on the style feature extraction.
We would like the network to learn appropriate style representations purely driven by the data.
In fact, we have considered employing other common losses, but found the results have little changes,
which proves that our objective design has been robust enough to learn a smart way of stylization from the data.

% 我们并没有施加其他的约束希望网络能在数据的驱动下学会字效特征的表示，
% 在实验中，我们考虑加入常见的loss，但是发现并没有带来明显的改进
% 我们认为，加入过多的因素可能反而具有负面效果，
% 这说明我们的网络设计使得网络根据数据学会到足够聪明的风格化方式。。。

\subsection{One-Shot Text Effects Transfer}
\label{sec:fewshot}

Learning-based methods are heavily dependent on dataset by nature and usually require thousands of images for training. We have collected a text effects dataset, and our TET-GAN can be well trained using this dataset to generate as much as $64$ different text effects.
However, it is still quite expensive with respect to date collection of user-customized styles. To develop a system that supports personalized text effects transfer,
we build upon our well-trained TET-GAN, and propose a novel ``self-stylization'' training scheme for one-shot learning, where only one training pair is required. Furthermore, we show that our network can be extended to solve a more challenging unsupervised problem where only one example style image is available.

\begin{figure}[t]
  \centering
  \includegraphics[width=0.96\linewidth]{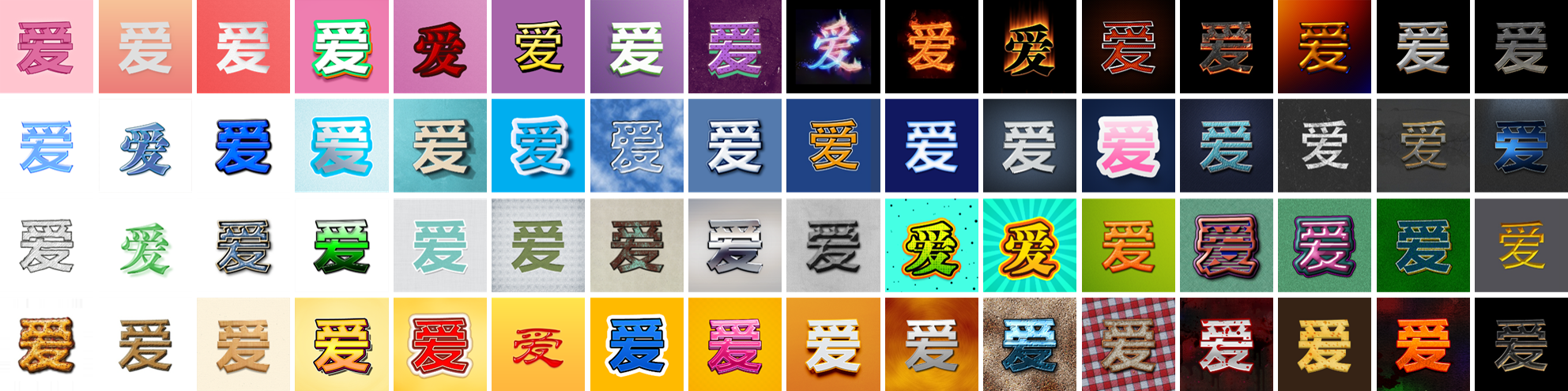}
  \caption{An overview of our text effects dataset.}\label{fig:dataset}
\end{figure}

\begin{figure}[t]
  \centering
  \subfigure[]{
  \includegraphics[width=0.31\linewidth]{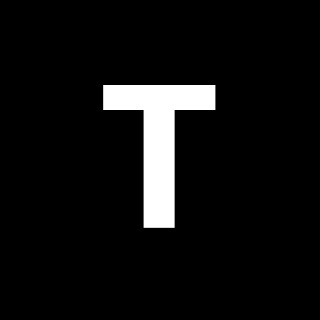}}
  \subfigure[]{
  \includegraphics[width=0.31\linewidth]{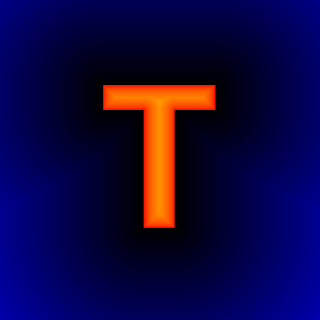}}
  \subfigure[]{
  \includegraphics[width=0.31\linewidth]{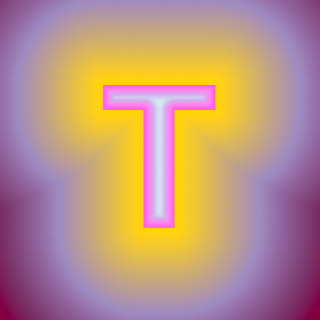}}
  \caption{Distribution-aware data augmentation. (a) Raw text image.
  (b) Result of distribution-aware text image preprocessing.
  (c) Result of distribution-aware text effects augmentation by tinting (b) using random colormaps.}\label{fig:preprocessing}
\end{figure}

% 如图XX所示，训练好的网络已经学会了关键的字效根据字形重构的方法。
% 对于新的用户指定的字效，网络只需要额外重点学习如何刻画纹理细节。
% 为此，我们提出了一个简单而高效的自风格化策略，
\textbf{One-shot supervised learning}. As shown in Fig.~\ref{fig:finetune}(b), for an unseen user-specified style, the network trained on our dataset has learned to generate the basic structure of the text effects.
%It only needs to focus on the texture details.
It only needs to be finetuned to better reconstruct the texture details of the specified style.
To achieve this goal, we propose a simple and efficient ``self-stylization'' training scheme.
Specifically, as shown in Fig.~\ref{fig:finetune}(d), we randomly crop the images to obtain a bunch of text effects images that have the same style but differ in the pixel domain. They constitute a training set to finetune our network to generate vivid textures as shown in Fig.~\ref{fig:finetune}(c).

\begin{figure*}[h]
\centering
\includegraphics[width=1\linewidth]{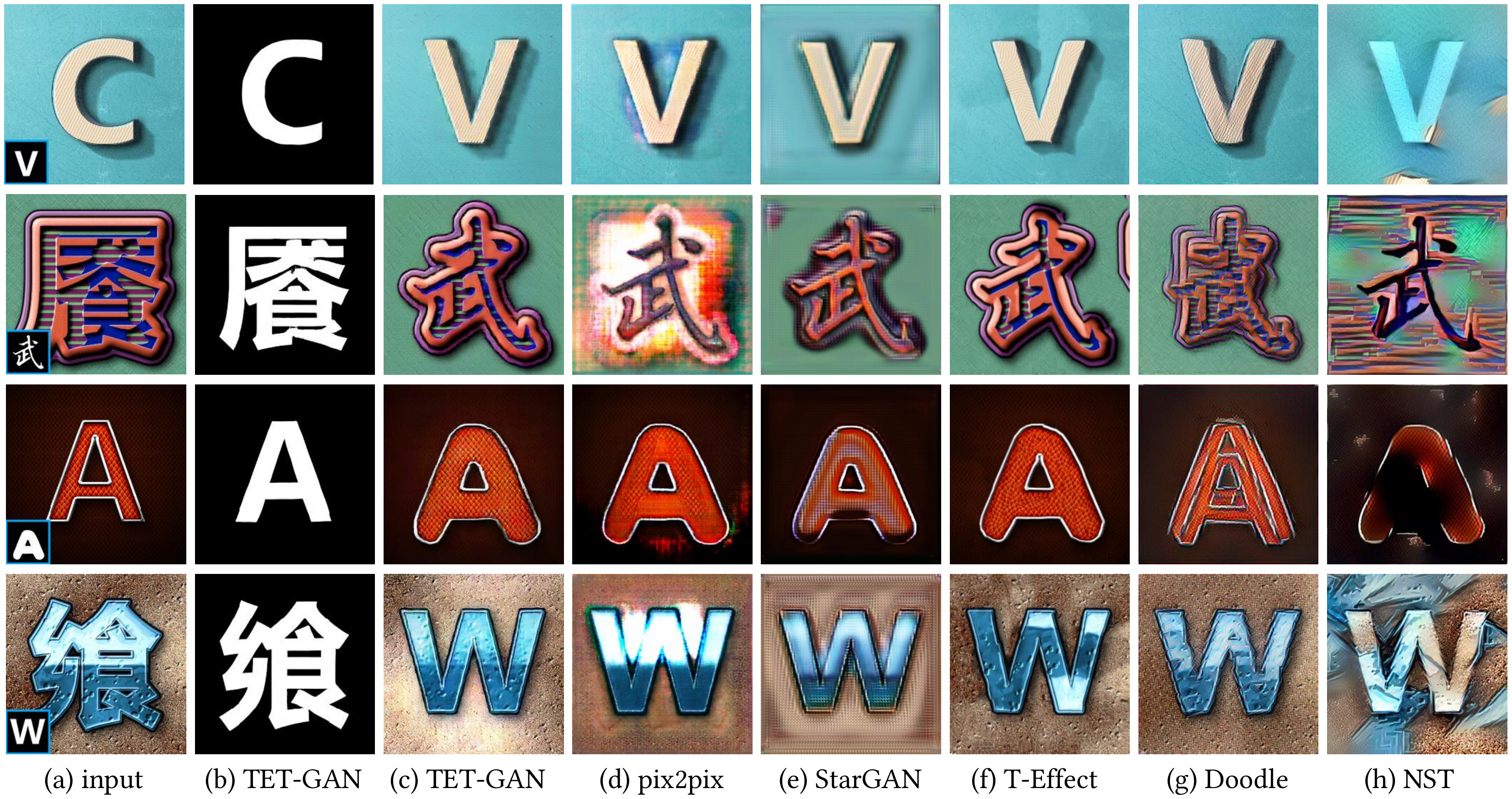}
\caption{Comparison with state-of-the-art methods on various text effects. (a) Input example text effects with the target text in the lower-left corner. (b) Our destylization results. (c) Our stylization results. (d) pix2pix-cGAN~\cite{Isola2017Image}.  (e) StarGAN~\cite{Choi2017StarGAN}. (f) T-Effect~\cite{Yang2017Awesome}. (g) Neural Doodles~\cite{Champandard2016Semantic}. (h) Neural Style Transfer~\cite{gatys2016image}. }
\label{fig:comparison}
\end{figure*}

% 然而，将x~视为gt的做法使得content feature缺乏足够精确的目标，
\textbf{One-shot unsupervised learning}. Our network architecture gives us great flexibility. It is intuitive to exploit the destylization subnetwork to generate the text image from the new text effects image, and use this image pair for one-shot supervised learning. In other word, $\tilde{x}=G_{\mathcal{X}}(E^{c}_{\mathcal{Y}}(y))$ is used as an auxiliary $x$ during the finetuning.
However, the accuracy of $\tilde{x}$ cannot be guaranteed, which may mislead the extraction of content features.
To solve this problem, a style autoencoder reconstruction loss is employed, which further constrains the content features to reconstruct the input text effects image with the style features:
\begin{equation}\label{eq:unsuper_loss}
  \mathcal{L}_{\text{srec}}=\lambda_{\text{srec}}\mathbb{E}_{y}[\|G_{\mathcal{Y}}(E^{c}_{\mathcal{Y}}(y),E^{s}_{\mathcal{Y}}(y))-y\|_{1}].
\end{equation}
And our objective for unsupervised learning takes the form
\begin{equation}\label{eq:unsuper_loss}
  \min_{E,G}\max_{D} \mathcal{L}_{\text{rec}}+\mathcal{L}_{\text{desty}}+\mathcal{L}_{\text{sty}}+\mathcal{L}_{\text{srec}}.
\end{equation}

\section{Distribution-Aware Data Collection and Augmentation}

% 秀一张overview，一张文字图片变成的距离距离相关的。。。，一张根据那个生成的字效图
We propose a new dataset including $64$ text effects each with $775$ Chinese characters, $52$ English letters and $10$ Arabic numerals, where the first $708$ Chinese characters are for training and others for testing. Fig.~\ref{fig:dataset} shows an overview of these text effects. Each text effects image has a size of $320\times320$ and is provided with its corresponding text image.
Our dataset contains two text effects kindly provided by the authors of~\cite{Yang2017Awesome}.
To generate other $62$ text effects, we first collected psd files released by several text effects websites, or created psd files ourselves following the tutorials on these websites. Then we used batch tools and scripts to
automatically replace characters and produced $837$ text effects images for each psd file.

\textbf{Distribution-aware text image preprocessing}. As reported in~\cite{Yang2017Awesome}, the spatial distribution of the texture in text effects is highly related to its distance from the glyph, forming an effective prior for text effects transfer. To leverage this prior, we propose a distribution-aware preprocessing for the text image to directly feed our network with distance cues. As shown in Fig.~\ref{fig:preprocessing}, we extend the raw text image from one channel to three channels. The R channel is the original text image, while G channel and B channel are distance maps where the value of each pixel is its distance to the background black region and the foreground white glyph, respectively. Another advantage of the preprocessing is that our three-channel text images have much fewer saturated areas than the original ones, which greatly facilitates the extraction of valid features.

\textbf{Distribution-aware text effects augmentation}. Besides the text images, we further propose the distribution-aware augmentation of the text effects images. The key idea is to augment our training data by generating random text effects based on the pixel distance from the glyph. Specifically, we first establish a random colormap for each of the R and G channels, which maps each distance value to a corresponding color. Then we use the colormaps of the R and G channels to tint the background black region and the foreground white glyph in the text image separately.
Fig.~\ref{fig:preprocessing}(c) shows an example of the randomly generated text effects images. These images whose colors are distributed strictly according to distance can effectively guide our network to discover the spatial relationship between the text effects and the glyph.
In addition, data augmentation could also increase the generalization capabilities of the network.

\section{Experimental Results}

\subsection{Implementation Details}

\textbf{Network architecture}. We adapt our network architectures from pix2pix-cGAN~\cite{Isola2017Image}.
The three encoders use a same structure built with Convolution-BatchNorm-ReLU layers,
and the decoders are built with Deconvolution-BatchNorm-LeakyReLU layers.
The architecture of our two discriminators follows PatchGAN~\cite{Isola2017Image}.
We add skip connections between the sharing layers of encoders and decoders so that they form a UNet~\cite{Ronneberger2015U}. By doing so, our network can capture both low-level and high-level features.
Considering Instance Normalization (IN)~\cite{Ulyanov2017Instance} can better characterize the style of each image instance for a robust style removal than Batch Normalization (BN)~\cite{Ioffe2015Batch}, we further replace BN with IN in $E^{c}_{\mathcal{Y}}$, which can effectively improve the destylization results.

% Tero Karras, Timo Aila, Samuli Laine, and Jaakko Lehtinen. 2017. Progressive growing of gans for improved quality, stability, and variation. arXiv preprint arXiv:1710.10196 (2017).
% ICLR18 Progressive Growing of GANs for Improved Quality, Stability, and Variation
\textbf{Network training}. We train the network on the proposed dataset. All images are cropped to $256\times256$ and one quarter of the training samples use the augmented text effects.
To stabilize the training of GAN, we follow the very recent works of progressive training strategies~\cite{Karras2017Progressive}.
The inner layers of our generators are first trained on downsampled $64\times64$ images. Outer layers are then added progressively to increase the resolution of the generated images until the original resolution is reached. When new layers are added to the encoders, we fade them in smoothly to avoid drastic network changes~\cite{Karras2017Progressive}.
Adam optimizer is applied with a fixed learning rate of $0.0002$ and a batch size of $32, 16$ and $8$ for image size of $64\times64, 128\times128$ and $256\times256$, respectively. For all experiments, we set $\lambda_{\text{dfeat}}=\lambda_{\text{dpix}}=\lambda_{\text{spix}}=\lambda_{\text{rec}}=\lambda_{\text{srec}}=100$, $\lambda_{gp}=10$, and
$\lambda_{\text{dadv}}=\lambda_{\text{sadv}}=1$.

\begin{figure}[t]
  \centering
  \includegraphics[width=0.98\linewidth]{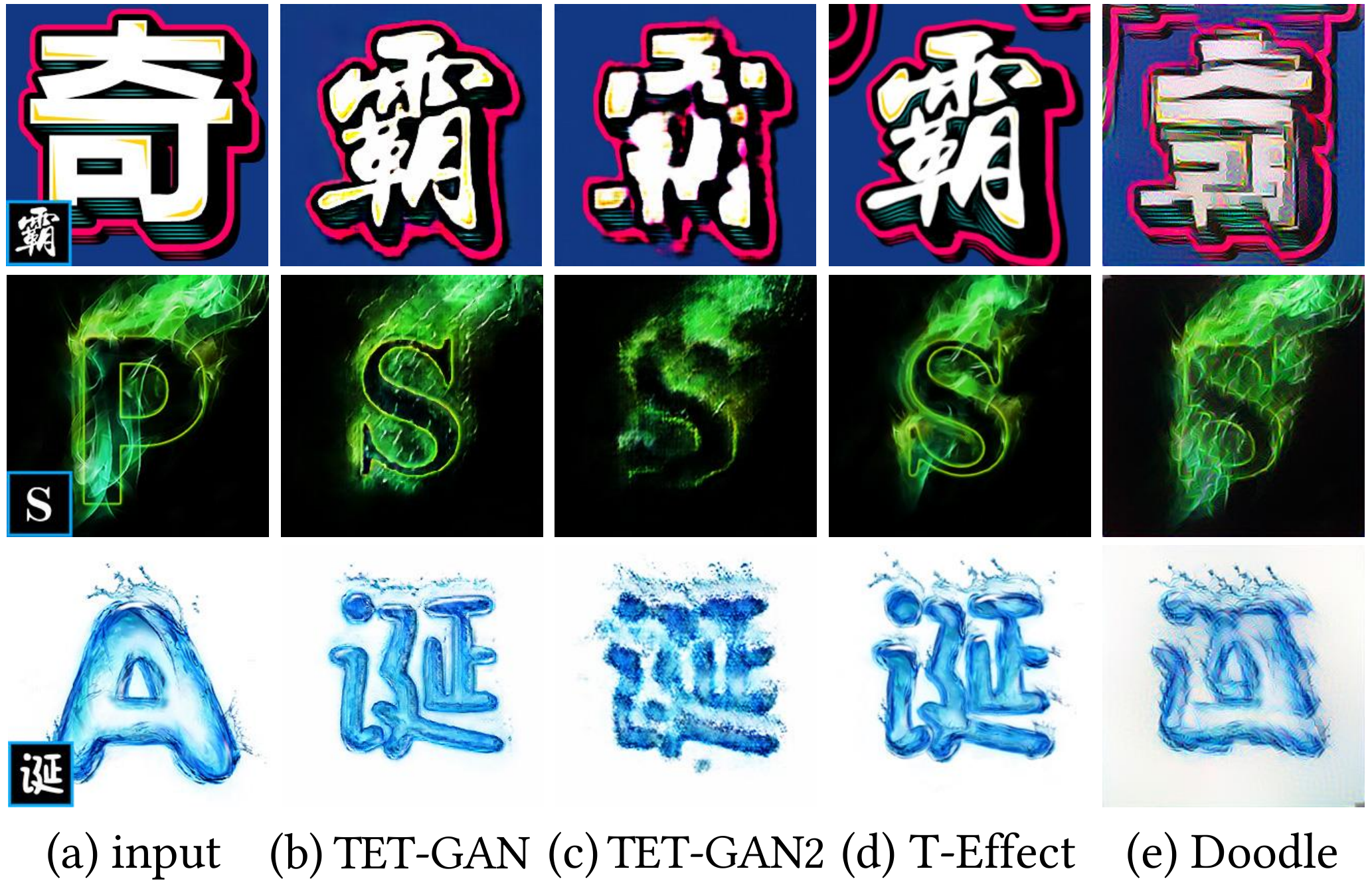}
  \caption{Comparison with other methods on one-shot supervised style transfer. (a) Example text effects and the target text. (b) Our results. (c) Results of our network without pretraining. (d) T-Effect~\cite{Yang2017Awesome}. (e) Neural Doodles~\cite{Champandard2016Semantic} }\label{fig:comparison2}
\end{figure}

\begin{figure}[t]
  \centering
  \includegraphics[width=0.98\linewidth]{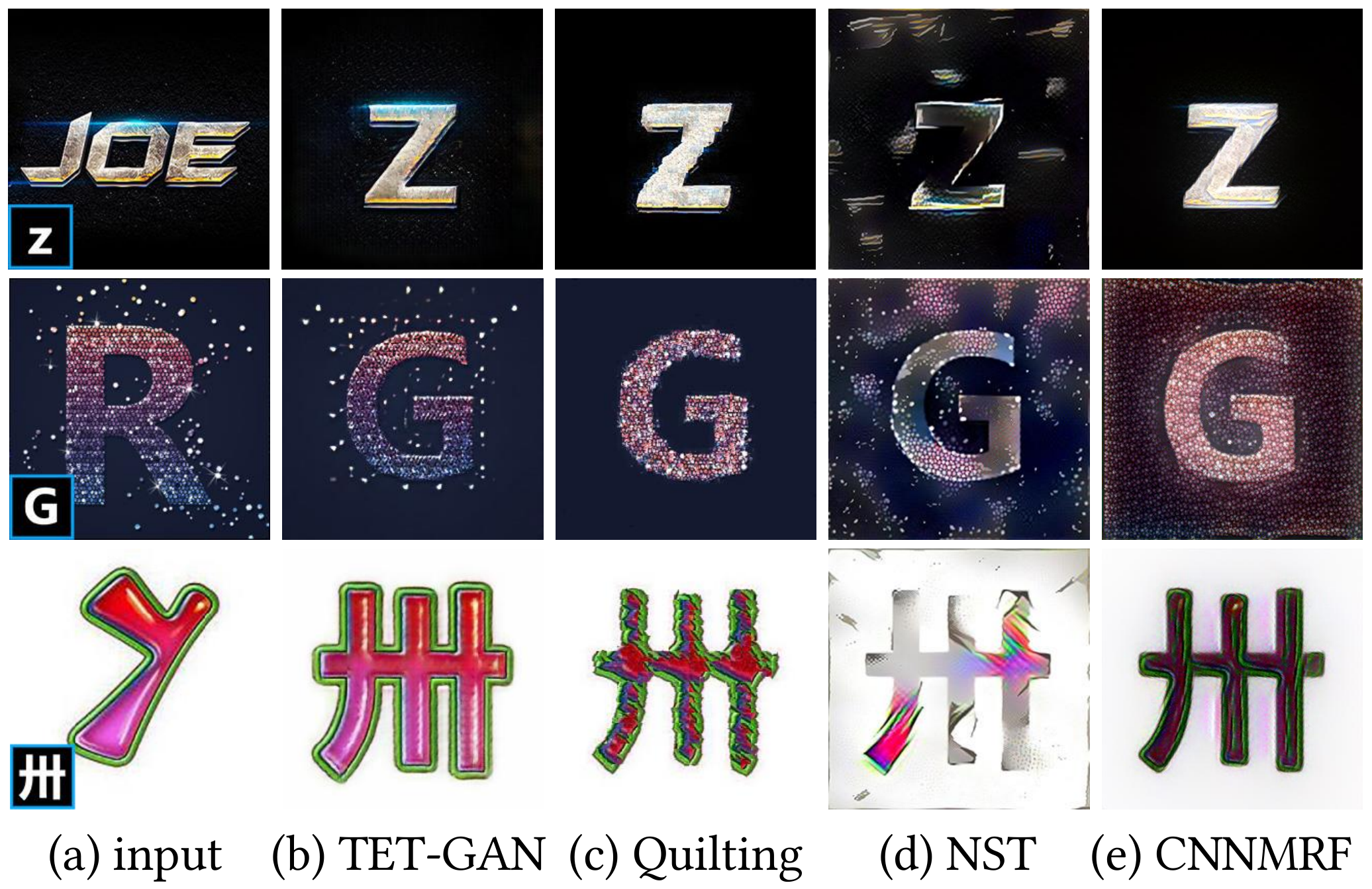}
  \caption{Comparison with other methods on one-shot unsupervised style transfer.
  (a) Example text effects and the target text. (b) Our results. (c) Image Quilting~\cite{Efros2001Image}. (d) Neural Style Transfer~\cite{gatys2016image}. (e) CNNMRF~\cite{Li2016Combining}}\label{fig:comparison3}
\end{figure}

\subsection{Comparison with State-of-the-Art}

In Fig.~\ref{fig:comparison}, we present a comparison of our network with five state-of-the-art style transfer methods.
The first two methods, pix2pix-cGAN and StarGAN, both employ GAN for domain transfer, and can be trained on our dataset to handle text effects. To order to allow pix2pix-cGAN to handle multiple styles, we change its input from a single text image to a concatenation of three images: the example text effects image, its corresponding glyph and the target glyph.
Pix2pix-cGAN fails to completely adapt the style image to the new glyphs, creating some ghosting artifacts. Meanwhile, the texture details are not fully inferred, leaving some flat or over-saturated regions.
StarGAN learns some color mappings, but fails to synthesize texture details and suffers from distinct checkerboard artifacts.
The following three methods are designed for zero-shot style transfer.
T-Effect and Neural Doodles synthesize textures using local patches under the glyph guidance of the example image. T-Effect processes patches in the pixel domain, leading to obvious color discontinuity. Instead, Neural Doodles uses deep-based patches for better patch fusion but fails to preserve the shape of the text.
Neural Style Transfer cannot correctly find the correspondence between the texture and text, creating interwoven textures.
By comparison, our network learns valid glyph features and style features, thus precisely transferring text effects with the glyph well protected.
We additionally show our destylization results in the second column, where the style features are effectively removed.

We compare our network with T-Effect and Neural Doodles on supervised stylization with only one observed example pair in Fig.~\ref{fig:comparison2}. Our method is superior to Neural Doodles in glyph preservation and is comparable to T-Effect. More importantly, in terms of efficiency, T-Effect takes about one minute per image, while our method only takes about $20$ms per image after a three-minute finetuning. In addition, as shown in Fig.~\ref{fig:comparison2}(c), if trained from scratch, the performance of our network drops dramatically, verifying that pretraining on our dataset successfully teaches our network the domain knowledge of text effects synthesis.
In Fig.~\ref{fig:comparison3}, we further compare with three methods on the challenging unsupervised stylization with only one observed example, where the advantages of our approach are more pronounced.

\begin{figure}[t]
  \centering
  \includegraphics[width=0.98\linewidth]{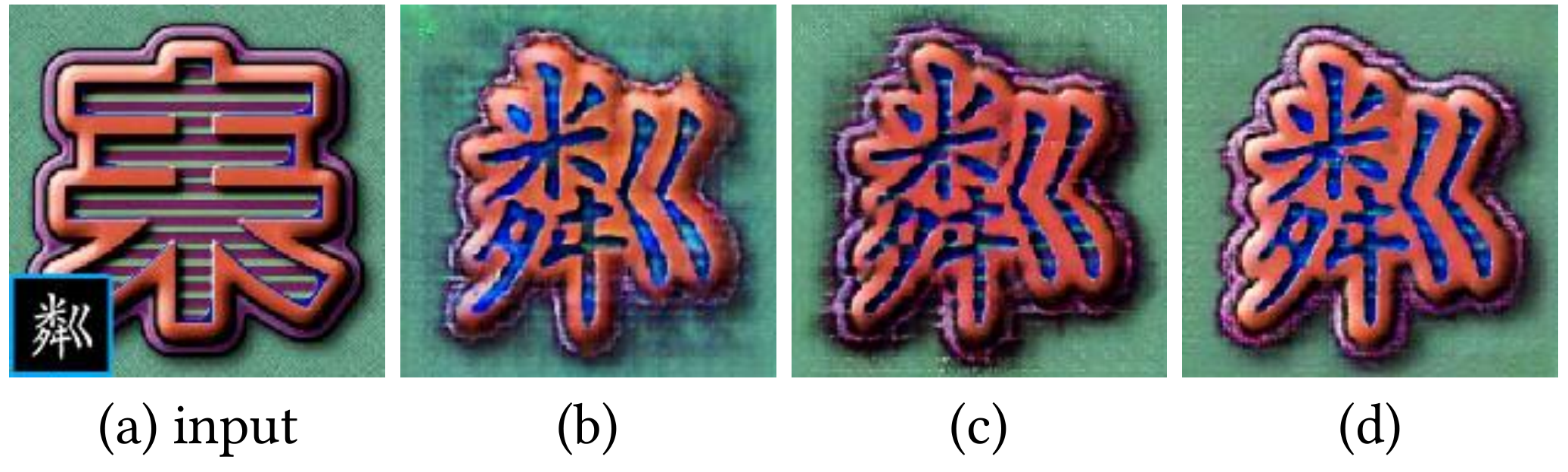}
  \caption{Effect of the reconstruction loss and feature loss. (a) Input. (b) Model without $\mathcal{L}_{\text{rec}}$ and $\mathcal{L}_{\text{dfeat}}$. (c) Model without $\mathcal{L}_{\text{dfeat}}$. (d) Full model.}\label{fig:ablation1}
\end{figure}

\begin{figure}[t]
  \centering
  \includegraphics[width=0.98\linewidth]{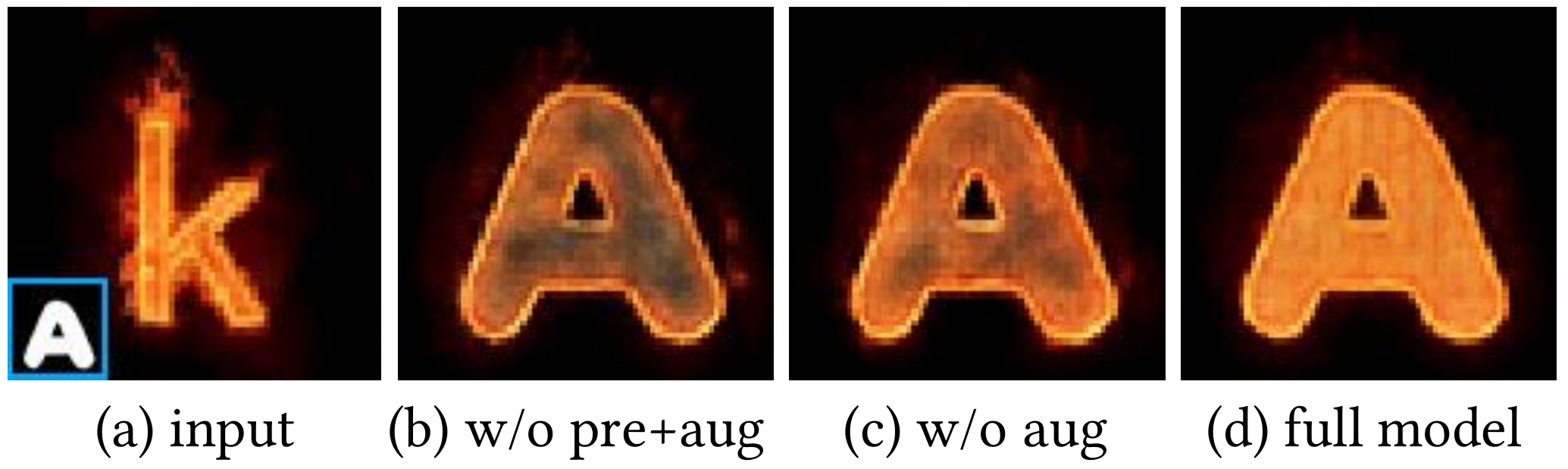}
  \caption{A comparison of results with and without our distribution-aware data preprocessing and augmentation.}\label{fig:ablation2}
\end{figure}

\subsection{Ablation Study}

% We perform the ablation experiment to study the role of each part in CartoonGAN.
In Fig.~\ref{fig:ablation1}, we study the effect of the reconstruction loss (Eq.~(\ref{eq:desty_feat_loss})) and the feature loss (Eq.~(\ref{eq:ae_rec_loss})). Without these two losses, even the color palette of the example style is not correctly transferred. In Fig.~\ref{fig:ablation1}(c), the glyph is not fully disentangled from the style, leading to annoying bleeding artifacts. The satisfying results in Fig.~\ref{fig:ablation1}(d) verify that our feature loss effectively guides TET-GAN to extract valid content representations to synthesize clean text effects.

In Fig.~\ref{fig:ablation2}, we examine the effect of our distribution-aware text image preprocessing and text effects augmentation through a comparative experiment. %The target `A' is much bolder than the  example `D'.
Without the preprocessing and augmentation, the inner flame textures are not synthesized correctly.
As can be seen in Fig.~\ref{fig:ablation2}(d), our distribution-aware data augmentation strategy helps the network learn to infer textures based on their correlated position on the glyph, and thus the problem is well solved.

\begin{figure}[t]
  \centering
  \includegraphics[width=0.96\linewidth]{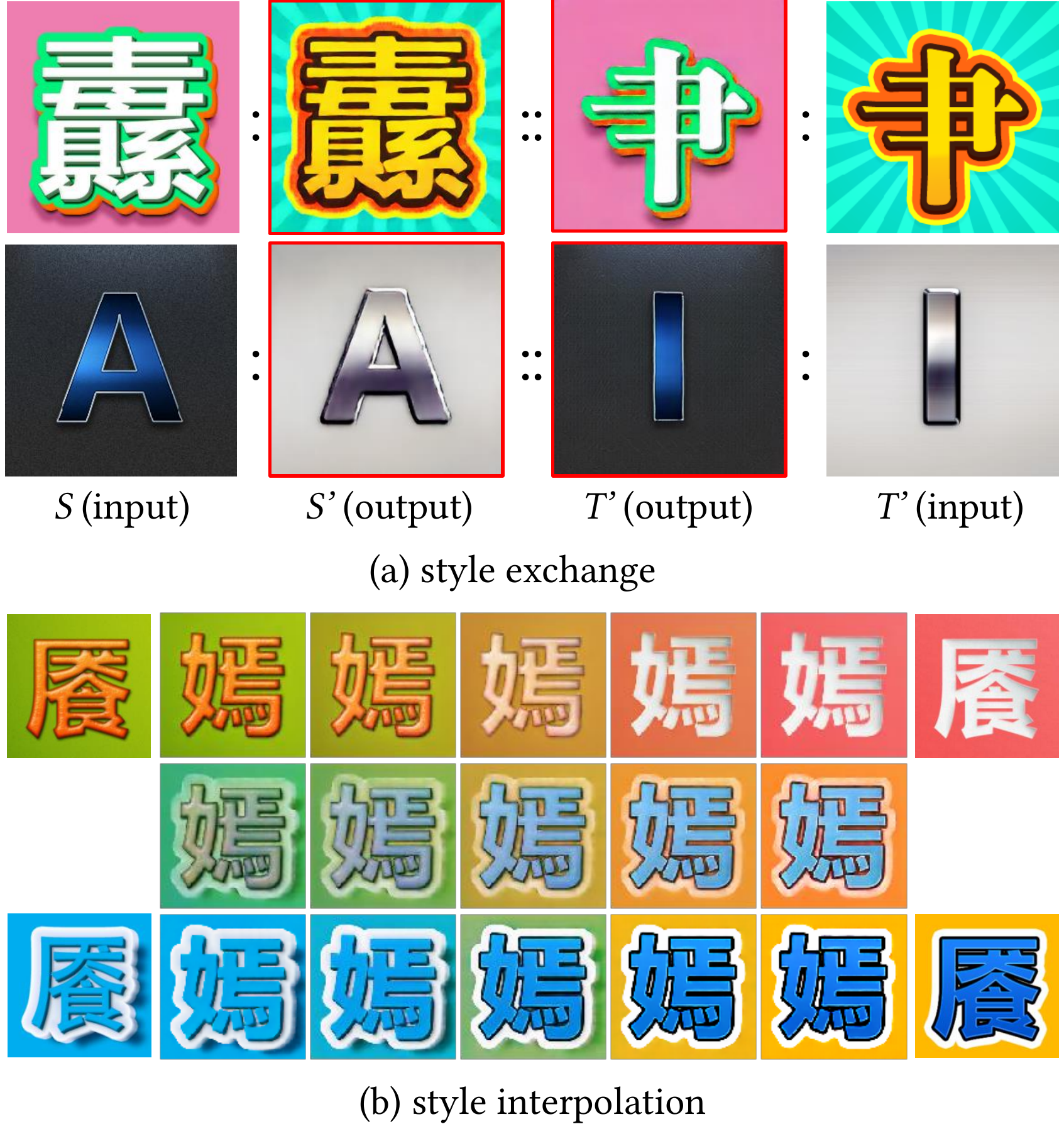}
  \caption{Applications of TET-GAN.}\label{fig:app}
\end{figure}

\subsection{Application}

The flexibility of TET-GAN is further manifested by two applications: style exchange and style interpolation. First, we can exchange the styles from two text effects images as shown in Fig.~\ref{fig:app}(a). It is accomplished by extracting the glyphs using the destylization subnetwork and then applying the styles to each other using the stylization subnetwork.
Second, the explicit style representations enable intelligent style editing. Fig.~\ref{fig:app}(b) shows an example of style fusion. We interpolate between four different style features, and decode the integrated features back to the image space, obtaining brand-new text effects.

\begin{figure}[t]
  \centering
  \includegraphics[width=0.98\linewidth]{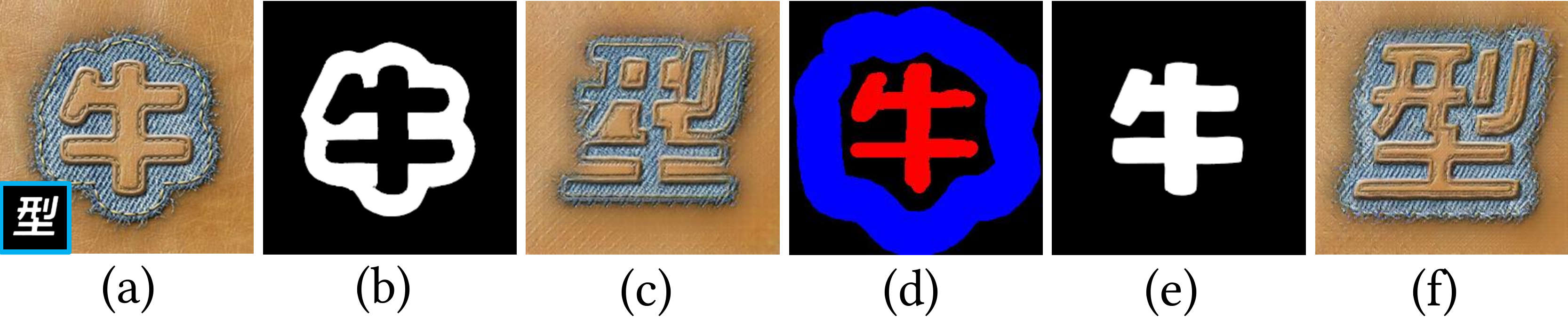}
  \caption{User-interactive unsupervised style transfer. (a) Example text effects and the target text. (b)(c) Our destylization and stylization results after finetuning. (d) A mask provided by the user, where the blue and red regions indicate the background and foreground, respectively. (e)(f) Our destylization and stylization results with the help of the mask.  }\label{fig:fail}
\end{figure}

\subsection{Failure Case}

While our approach has generated appealing results, some limitations still exist.
Our destylization subnetwork is not fool-proof due to the extreme diversity of the text effects, which may totally differ from our collected text effects.
Fig.~\ref{fig:fail} shows a failure case of one-shot unsupervised text effects transfer.
Our network fails to recognize the glyph. As a result, in the stylization result, the text effects in the foreground and background are reversed.
This problem can be possibly solved by user interaction. Users can simply paint a few strokes (Fig.~\ref{fig:fail}(d)) to provide a priori information about the foreground and the background, which is then fed into the network as a guidance to constrain the glyph extraction and thereby improve the style transfer results (Fig.~\ref{fig:fail}(f)).

\section{Conclusion}

In this paper, we present a novel TET-GAN for text effects transfer.
We integrate stylization and destylization into one uniform framework to jointly learn valid content and style representations of the artistic text.
Exploiting explicit style and content representations, TET-GAN is able to transfer, remove and edit dozens of styles, and can be easily customized with user-specified text effects.
In addition, we develop a dataset of professionally designed text effects to facilitate researches.
Experimental results demonstrate the superiority of TET-GAN in generating high-quality artistic typography.
As a future direction, one may explore other more sophisticated style editing methods, such as background replacement, color adjustment and texture attribute editing.

\bibliographystyle{aaai}
\bibliography{egbib}

\end{document}